\documentclass[11pt]{article}
\usepackage{coling2020}
\usepackage{times}
\usepackage{url}
\usepackage{latexsym}
\usepackage{graphicx}

\title{Measuring the Novelty of Natural Language Text Using the\\ Conjunctive Clauses of a Tsetlin Machine Text Classifier}

\author{Bimal Bhattarai \\
    University of Agder \\
    Grimstad, Norway\\
  {\tt bimal.bhattarai@uia.no} \\
  \And
    Ole-Christoffer Granmo \\
    University of Agder  \\
    Grimstad, Norway\\
  {\tt ole.granmo@uia.no}
  \And
    Lei Jiao \\
    University of Agder  \\
    Grimstad, Norway\\
  {\tt  lei.jiao@uia.no}}

\date{}
\begin{document}

\maketitle
\begin{abstract}
  Most supervised text classification approaches assume a closed world, counting on all classes being present in the data at training time. This assumption can lead to unpredictable behaviour during operation, whenever novel, previously unseen, classes appear. Although deep learning-based methods have recently been used for novelty detection, they are challenging to interpret due to their black-box nature. This paper addresses \emph{interpretable} open-world text classification, where the trained classifier must deal with novel classes during operation. To this end, we extend the recently introduced Tsetlin machine (TM) with a novelty scoring mechanism. The mechanism uses the conjunctive clauses of the TM to measure to what degree a text matches the classes covered by the training data. We demonstrate that the clauses provide a succinct interpretable description of known topics, and that our scoring mechanism makes it possible to discern novel topics from the known ones. Empirically, our TM-based approach outperforms seven other novelty detection schemes on three out of five datasets, and performs second and third best on the remaining, with the added benefit of an interpretable propositional logic-based representation.
\end{abstract}

\section{Introduction}

In recent years, deep learning-based techniques have achieved superior performance on many text classification tasks. Most of the classifiers use supervised learning, assuming a closed-world environment \cite{2}. That is, the classes present in the test data (or during operation) are also assumed to be present in the training data.  However, when facing an open-world environment, new classes may appear after training \cite{1}.  In such cases, assuming a closed world can lead to unpredictable behaviour. For example, a chatbot interacting with a human user will regularly face new user intents that it has not been trained to recognize. A chatbot for banking services may, for instance, have been trained to recognize the intent of applying for a loan. However, it will provide meaningless responses if it fails to recognize that asking for a lower interest rate is a new and different user intent. The problem with neural network-based supervised classifiers that use the typical softmax layer is that they erroneously force novel input into one of the previously seen classes, by normalizing the class output scores to produce a distribution that sums to $1.0$. Instead, a robust classifier should be able to flag input as novel, rejecting to label it according to the presently known classes. Recently, many important application areas make use of novelty detection such as medical applications, fraud detection \cite{12}, sensor networks \cite{13} and text analysis \cite{14}. For a further study of these classes of techniques, the reader is referred to \cite{11}.

\begin{figure}[ht]
    \centering
    \includegraphics[width=15cm]{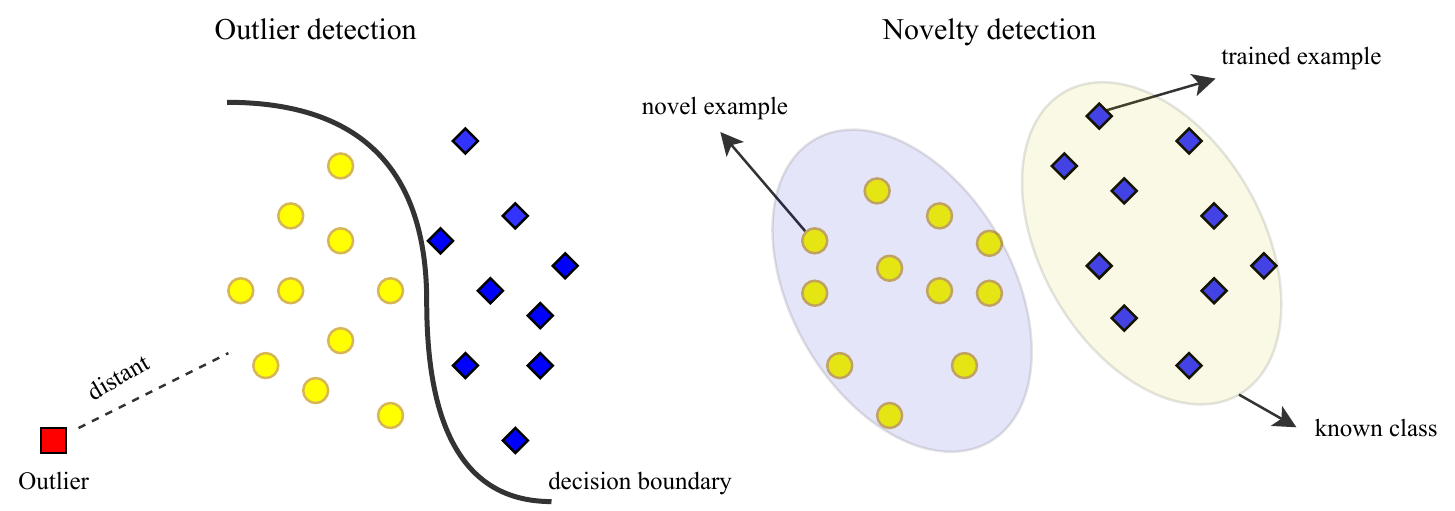}
    \caption{Visualization of outlier detection and novelty detection.}
    \label{fig:novelty_vizualization}
\end{figure}
\par
\par
 Figure \ref{fig:novelty_vizualization} illustrates the problem of novelty detection, i.e., recognizing when the data fed to a classifier is novel and somehow differs from the data that was available during training.
In brief, after training on data with known classes (blue data points to the right in Figure \ref{fig:novelty_vizualization}), the classifier should be able to detect novel data arising from new, previously unseen, classes (yellow data points to the right in the figure). We will refer to the known data points as \emph{positive examples} and the novel data points as \emph{negative examples}. Note that the problem of novelty detection is closely related to so-called outlier detection \cite{4}\cite{18}. However, as exemplified to the left in Figure \ref{fig:novelty_vizualization}, the latter problem involves flagging data points that are part of an already known class, yet deviating from the other data points, e.g., due to measurement errors or anomalies (red point).

\textbf{Problem Definition:} Generally, in multiclass classification, we have a set of example data points \(X=(x_i, y_i )\), \(x_i\) \(\in R^s,\) where \(i=1,2,3,\ldots,N\) indexes the positive examples, \(s\) is the dimensionality of the data point \(x_i\), and \(y_i\) is the label of \(x_i\), assigning it a class. For any data point \(x_i\), also referred to as a feature vector, a classifier function \(\hat{y}=f(x; X)\) is to assign the data point a predicted class $\hat{y}$, after the function has been fitted to the training data $X$.
Additionally, a novelty scoring function $z(x; X)$, also fitted on $X$, calculates a novelty score, so that a threshold $\sigma$ can be used to recognize novel input. In other words, the classifier is to return the correct class label while simultaneously being capable of rejecting novel examples. 
\par
Under the above circumstances, standard supervised learning would fail, particularly for methods based on building a discriminant function, such as neural networks. As shown to the left of Figure \ref{fig:novelty_vizualization}, a discriminant function captures the discriminating ``boundaries" between the known classes and cannot readily be used to discern novel classes. Still, a traditional way of implementing novelty detection is to threshold the entropy of the class probability distribution \cite{6}. Such methods are not actually measuring novelty, but closeness to the decision boundary. Such an approach thus leads to undetected novel data points when the data points are located far from the decision boundary. Another, and perhaps more robust, approach is to take advantage of the class likelihood function, which estimates the probability of the data given the class. 

\textbf{Paper Contributions:} Unlike traditional methods, in this paper we leverage the conjunctive clauses in propositional logic that a TM builds. TM clauses represent frequent patterns in the data, and our hypothesis is that these frequent patterns characterize the known classes succinctly and comprehensively. We thus establish a novelty scoring mechanism that is based simply on counting how many clauses match the input. This score can, in turn, be thresholded manually to flag novel input. However, for more robust novelty detection, we train several standard machine learning techniques to find accurate thresholds. The main contributions of our work can thus be summarized as follows:
\begin{itemize}
  \item We devise the first TM-based approach for novelty detection, leveraging the clauses of the TM architecture.
  \item We illustrate how the new technique can be used to detect novel topics in text, and compare the technique against widely used approaches on five different datasets.
\end{itemize}

\section{Related Work}


\par
The perhaps most common approach to novelty detection is distance-based methods \cite{22}, which assumes that the known or seen data are clustered together while novel data have a high distance to the clusters. The major drawback of these methods is computational complexity when performing clustering or nearest neighbor search in a large dataset. Early work on novelty detection also includes one-class SVMs \cite{8}, which are only capable of using the positive training examples to maximize the class margin. This shortcoming is overcome by the Center-Based Similarity (CBS) space learning method \cite{10}, which uses binary classifiers over vector similarities of training examples transformed into the center of the class. To build a classifier for detecting novel class distributions, Chow et al. proposed a confidence score-based method that suggests an optimum input rejection rule~\cite{15}. The method is relatively accurate, but it does not scale well to high-dimensional datasets.

\par
The novelty detection method OpenMax \cite{1} is more recent, and estimates the probability of the input belonging to a novel class. To achieve this, the method employs an extra layer, connected to the penultimate layer of the original network. The computational complexity of the method is however high, and the underlying inference cannot easily be interpreted for quality assurance. Lately, Yu et al. \cite{20} adopted the Adversarial Sample Generation (ASG) framework \cite{22} to generate positive and negative examples in an unsupervised manner. Then, based on those examples, they trained an SVM classifier for novelty detection. Furthermore, in computer vision, Scheirer et al. introduced the concept of open space risk to recognize novel image content \cite{2}. They proposed a “1-vs-set machine” that creates a decision space using a binary SVM classifier, with two parallel hyperplanes bounding the non-novel regions. In Section \ref{sec4}, we compare the performance of our new TM-based approach with the most widely used approaches among those mentioned above.


\section{Tsetlin Machine-based Novelty Detection}
In this section, we propose our approach to novelty detection based on the TM. First, we explain the architecture of the TM. Then, we show how novelty scores can be obtained from the TM clauses. Finally, we integrate the TM with a rule-based classifier for novel text classification.

\subsection{Tsetlin Machine (TM) Architecture}

The TM is a recent approach to pattern classification \cite{25} and regression \cite{26}. It builds on a classic learning mechanism called a Tsetlin automaton (TA), developed by M. L. Tsetlin in the early 1960s \cite{24}. In all brevity, multiple teams of TA combine to form the TM. Each team is responsible for capturing a frequent pattern of high precision, by composing a conjunctive clause. In-built resource allocation principles guide the teams to distribute themselves across the underlying sup-patterns of the problem. Recently, the TM has performed competitively with the state-of-the-art techniques, including deep neural networks for text classification \cite{27}. Further, the convergence of TM learning has been studied in \cite{zhang2020convergence}. In the following, we propose a new scheme that extends the TM with the capability of recognizing novel patterns.
\par

\begin{figure}[ht]
    \centering
    \includegraphics[width=14cm]{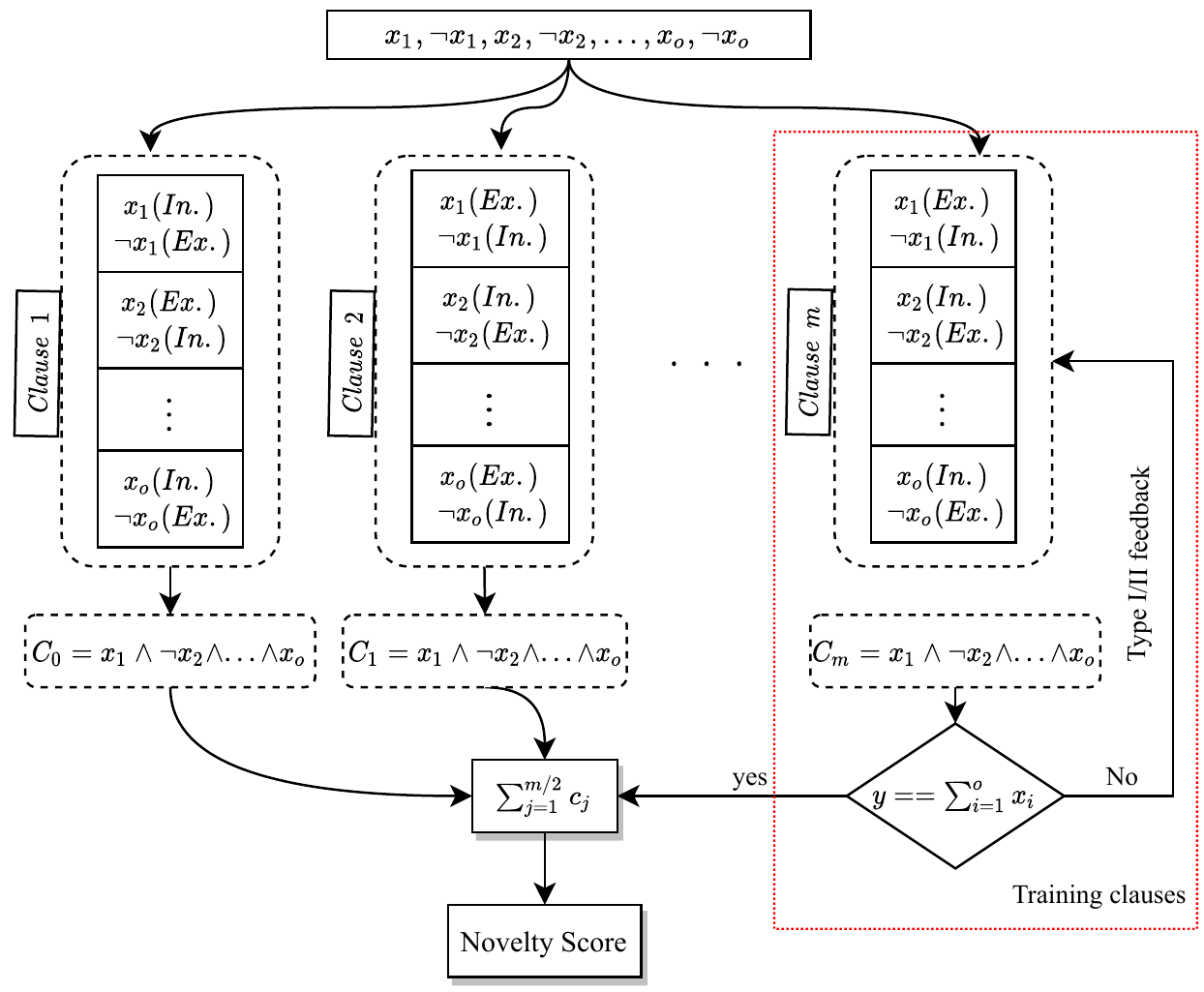}
    \caption{Optimization of clauses in TM for generating novelty score.}
    \label{fig:clauses_training}
\end{figure}

As explored in the following, Figure \ref{fig:clauses_training} capture the building blocks of a TM. As seen, a vanilla TM takes a vector $X=(x_1,\ldots,x_o)$ of binary features as input (Figure \ref{fig:tsetlin_framework}). We binarize text by using binary features that capture the presence/absence of terms in a vocabulary, akin to a bag of words, as done in \cite{27}. However, as opposed to a vanilla TM, our scheme does not output the predicted class. Instead, it calculates a novelty score per class. 

Together with their negated counterparts, $\bar{x}_k = \lnot x_k = 1-x_k$, the features form a literal set $L = \{x_1,\ldots,x_o,\bar{x}_1,\ldots,\bar{x}_o\}$. A TM pattern is formulated as a conjunctive clause $C_j$, formed by ANDing a subset $L_j \subseteq L$ of the literal set:
\begin{equation}
    C_j (X)=\bigwedge_{l_k \in L_j} l_k = \prod_{l_k \in L_j} l_k.
\end{equation}
For example, the clause $C_j(X) = x_1 x_2$ consists of the literals $L_j = \{x_1, x_2\}$ and outputs $1$ if $x_1 = x_2 = 1$. The number of clauses employed is a user set parameter $m$. Half of the clauses are assigned a positive polarity and the other half are assigned a negative polarity. The output of a conjunctive clause is determined by evaluating it on the input literals. When a clause outputs $1$, this means that it has recognized a pattern in the input. Conversely, the clause outputs $0$ when no pattern is recognized. The clause outputs, in turn, are combined into a classification decision through summation and thresholding using the unit step function $u$: 
\begin{equation}
    \hat{y} = u\left(\sum_{j=1}^{m/2} C_j^+(X) - \sum_{j=1}^{m/2} C_j^-(X)\right).
\end{equation}
 That is, the classification is performed based on a majority vote, with the positive clauses voting for $y=1$ and the negative for $y=0$. The classifier $\hat{y} = u\left(x_1 \bar{x}_2 + \bar{x}_1 x_2 - x_1 x_2 - \bar{x}_1 \bar{x}_2\right)$, e.g., captures the XOR-relation.
 
 A clause is composed by a team of TA, each TA deciding to \emph{Exclude} or \emph{Include} a specific literal in the clause. The TA learns which literals to include based on reinforcement: Type I feedback is designed to produce frequent patterns, while Type II feedback increases the discriminating power of the patterns (see \cite{25} for details).
 
 \subsection{Novelty Detection Architecture}
 
\begin{figure}[ht]
    \centering
    \includegraphics[width=12cm]{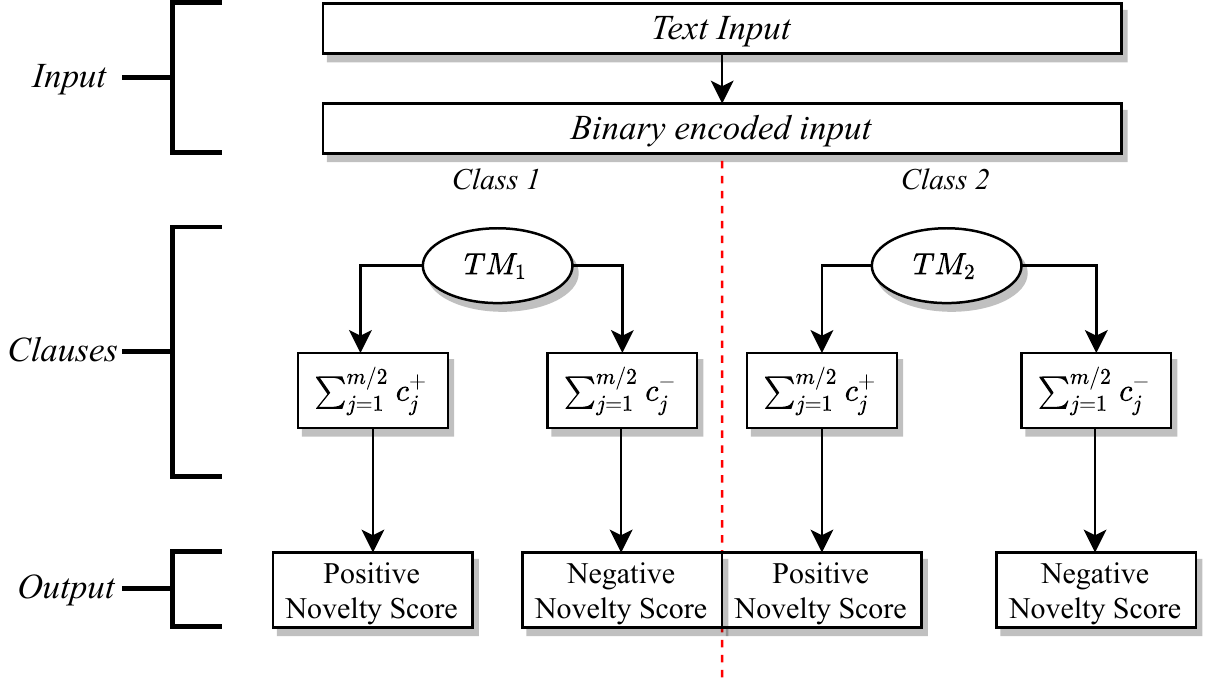}
    \caption{Multiclass Tsetlin Machine (MTM) framework to produce the novelty score for each class.}
    \label{fig:tsetlin_framework}
\end{figure}

 For novelty detection, however, we here propose to treat all clause output as positive, disregarding clause polarity. This is because both positive and negative clauses capture patterns in the training data, and thus can be used to detect novel input. We use this sum of absolute clause outputs as a novelty score, which denotes the resemblance of the input to the patterns formed by clauses during training. The resulting modified TM architecture is captured by Figure \ref{fig:tsetlin_framework}, showing how four different outputs are produced by the TM, two per class. These outputs form the basis for novelty detection.

\begin{figure}[ht]
    \centering
    \includegraphics[width=7cm]{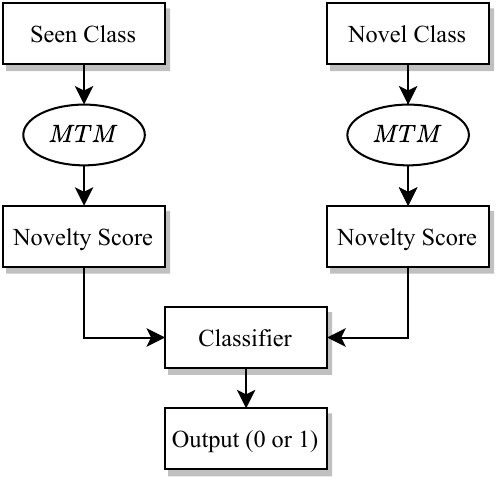}
    \caption{Novelty detection architecture.}
    \label{fig:system_architecture}
\end{figure}

The overall novelty detection architecture is shown in Figure \ref{fig:system_architecture}. Each TM (one per class) produces two novelty score for its respective class, one based on the positive polarity clauses and one based on the negative polarity ones. The novelty scores are normalized and then given to a classifier, such as decision tree (DT), K nearest neighbor (KNN), Support vector machine (SVM) and logistic regression (LR). The output from these classifiers are "Novel" or "Not Novel", i.e. \textit{0} or \textit{1} in the figure.

For illustration purposes, instead of using a machine learning algorithm to decide upon novelty, one can instead use a simple rule-based classifier, introducing a classification threshold $T$. Then the novelty score for input sentence can be compared with $T$ to detect whether a sentence is novel or not. That is, $T$ decides how many clauses must match to qualify the input as non-novel. The classification function $F(X)$ for a single class can accordingly be given as:

\[
F(X) = 
\left\{
\begin{array}{cl}
   1,  & \mathbf{if} \sum_{j=1}^{m/2}{C_j^+(X)} > T, \\
   1,  & \mathbf{if} \sum_{j=1}^{m/2}{C_j^-(X)} < -T, \\
   0,  & \mathbf{otherwise}.
\end{array}
\right.
\]

\section{Experiments and Results}\label{sec4}
In this section, we present the experimental evaluation of our proposed TM model and compare it with other baseline models across five datasets.

\subsection{Datasets}
We used the following datasets for evaluation.
\begin{itemize}
    \item 20-newsgroup dataset: The dataset contains 20 classes with a total of 18,828 documents. In our experiment, we consider the two classes \textit{``comp.graphics''} and \textit{``talk.politics.guns''} as known classes and the class \textit{``rec.sport.baseball''} as novel. We take 1000 samples from both known and novel classes for novelty classification.
    \item CMU Movie Summary corpus: The dataset contains 42,306 movie plot summaries extracted from Wikipedia and metadata extracted from Freebase. In our experiment, we consider the two movie categories \textit{``action''} and \textit{``horror''} as known classes and \textit{``fantasy''} as novel. 
    \item	Spooky Author Identification dataset: The dataset contains 3,000 public domain books from the following horror fiction authors: \textit{Edgar Allan Poe (EAP), HP Lovecraft (HPL)}, and \textit{Mary Shelley (MS)}. We train on written texts from \textit{EAP} and \textit{HPL} while treating texts from \textit{MS} as novel.
    \item	Web of Science dataset \cite{28}: This dataset contains 5,736 published papers, with eleven categories organized under three main categories. We use two of the main categories as known classes, and the third as a novel class.
    \item	BBC sports dataset: This dataset contains 737 documents from the BBC Sport website, organized in five sports article categorie and collected from 2004 to 2005. In our work, we use \textit{``cricket''} and \textit{``football''} as the known classes and \textit{``rugby''} as novel. 
\end{itemize}

\subsection{A Case Study}
To cast light on the interpretability of our scheme, we here use substrings from the 20 Newsgroup dataset, demonstrating novelty detection on a few simple cases. First, we form an indexed vocabulary set \(V\) including all literals from the dataset. The input text is binarized based upon the index of the literals in \(V\). For example, if a word in the text substring has been assigned index \(5\) in \(V\), the \(5^{th}\) position of the input vector is set to $1$. If a word is absent from the substring, its corresponding feature is set to $0$. Let us consider substrings from the two known classes and the novel class from the 20 Newsgroup dataset.
\begin{itemize}
    \item \textbf{Class} : comp.graphics (known)\\
    \textbf{Text}: Presentations are solicited on all aspects of Navy-related scientific visualization and virtual reality.\\
    \textbf{Literals}: \textit{``Presentations"}, \textit{``solicited"}, \textit{``aspects"}, \textit{``Navy"}, \textit{``related"}, \textit{``scientific"}, \textit{``visualization"}, \textit{``virtual"}, \textit{``reality"}.
    \item \textbf{Class}: talk.politics.guns (known)\\
    \textbf{Text}: Last year the US suffered almost 10,000 wrongful or accidental deaths by handguns alone. In the same year, the UK suffered 35 such deaths.\\
    \textbf{Literals}: \textit{``Last "}, \textit{``year"}, \textit{``US"}, \textit{``suffered"}, \textit{``wrongful"}, \textit{``accidental"}, \textit{``deaths"}, \textit{``handguns"}, \textit{``UK"}, \textit{``suffered"}.
    \item \textbf{Class} : rec.sport.baseball  (Novel)\\
    \textbf{Text}: The top 4 are the only true contenders in my mind. One of these 4 will definitely win the division unless it snows in Maryland.\\
    \textbf{Literals}: \textit{``top "}, \textit{``only"}, \textit{``true"}, \textit{``contenders"}, \textit{``mind"}, \textit{``win"}, \textit{``division"}, \textit{``unless"}, \textit{``snows"}, \textit{``Maryland"}
\end{itemize}
After training, the two known classes form conjunctive clauses that capture literal patterns reflecting the textual content. For the above example, we get the following clauses:
\begin{itemize}
    \item \(C_1^+ = \textit{``Presentations"} \wedge \textit{``aspects"}\wedge \textit{``Navy"}\wedge \textit{``scientific"}\wedge \textit{``virtual"}, \textit{``reality"}\wedge \textit{``year"}\wedge\textit{``US"}\\
    \wedge\textit{``mind"}\wedge\textit{``division"}\)
    \item \(C_1^- = \neg(\textit{``suffered"} \wedge \textit{``accidental"}\wedge \textit{``unless"}\wedge \textit{``snows"})\)
    \item \(C_2^+ = \textit{``last"} \wedge \textit{``year"}\wedge \textit{``US"}\wedge \textit{``wrongful"}\wedge \textit{``deaths"}, \textit{``accidental"}\wedge \textit{``handguns"}\wedge\textit{``Navy"}\\
    \wedge\textit{``snows"}\wedge\textit{``Maryland"}\wedge\textit{``divisions"}\)
    \item \(C_2^- = \neg(\textit{``presentations"} \wedge \textit{``solicited"}\wedge\textit{``virtual"}\wedge \textit{``top"}\wedge\textit{``win"})\)
\end{itemize}
Here, the clauses from each class captures the frequent patterns from the class. However, it may also contain certain literals from other classes. The positive polarity clauses provide evidence on the presence of a class, while negative polarity clauses provide evidence on the absence of the class. The novelty score for each class is calculated based on the propositional formula formed by the clauses. In general, for input from known classes,  the novelty scores are higher. This is because the clauses have been trained to vote for or against input from the known class. For example, when we pass a known class to our model, the clauses might produce scores as in Table \ref{novelty_score} (for illustration purposes).

\begin{table}[ht]
\begin{center}
\begin{tabular}{|c|c|c|c|c|c|c|c|}
\hline  Class & \(C_1^+\) &  \(C_1^-\)& \(C_2^+\) & \(C_2^-\) \\ \hline
Known & +6  &-3  &+3  & -5 \\ \hline
Novel & +2 & -1  & +1  &-2 \\
\hline
\end{tabular}
\end{center}
\caption{\label{novelty_score} Novelty score example when known text sentence is passed to the model. }
\end{table}

The scores are then used as features to prepare a dataset for employing machine learning classifiers to enhance novelty detection. As examplified in the table,  novelty scores for known classes are relatively higher than those of novel classes. This allows the final classifier to robustly recognize novel input, as explored empirically below.

\subsection{Empirical Evaluation}
We divided the task into two experiments i.e., 1) Novelty score calculation 2) Novelty/Known Class classification. In the first experiment, we employed the known classes to train the TM. The TM runs for 100 epochs with hyperparameter setting of 5000 clauses, a threshold \(T\) of 25, and a sensitivity \(s\) of 15.0.  Then, we use the clauses formed by the trained TM model to calculate the novelty scores for both the known and novel classes. We adopt an equal number of examples to gather the novelty score from both known and novel classes. In the second experiment, the novelty score generated from the first experiment is forwarded as input to standard machine learning classifiers, such as DT, KNN, SVM, LR, NB, and MLP to classify whether a text is novel. \par

\begin{figure}[ht]
    \centering
    \includegraphics[width=16cm]{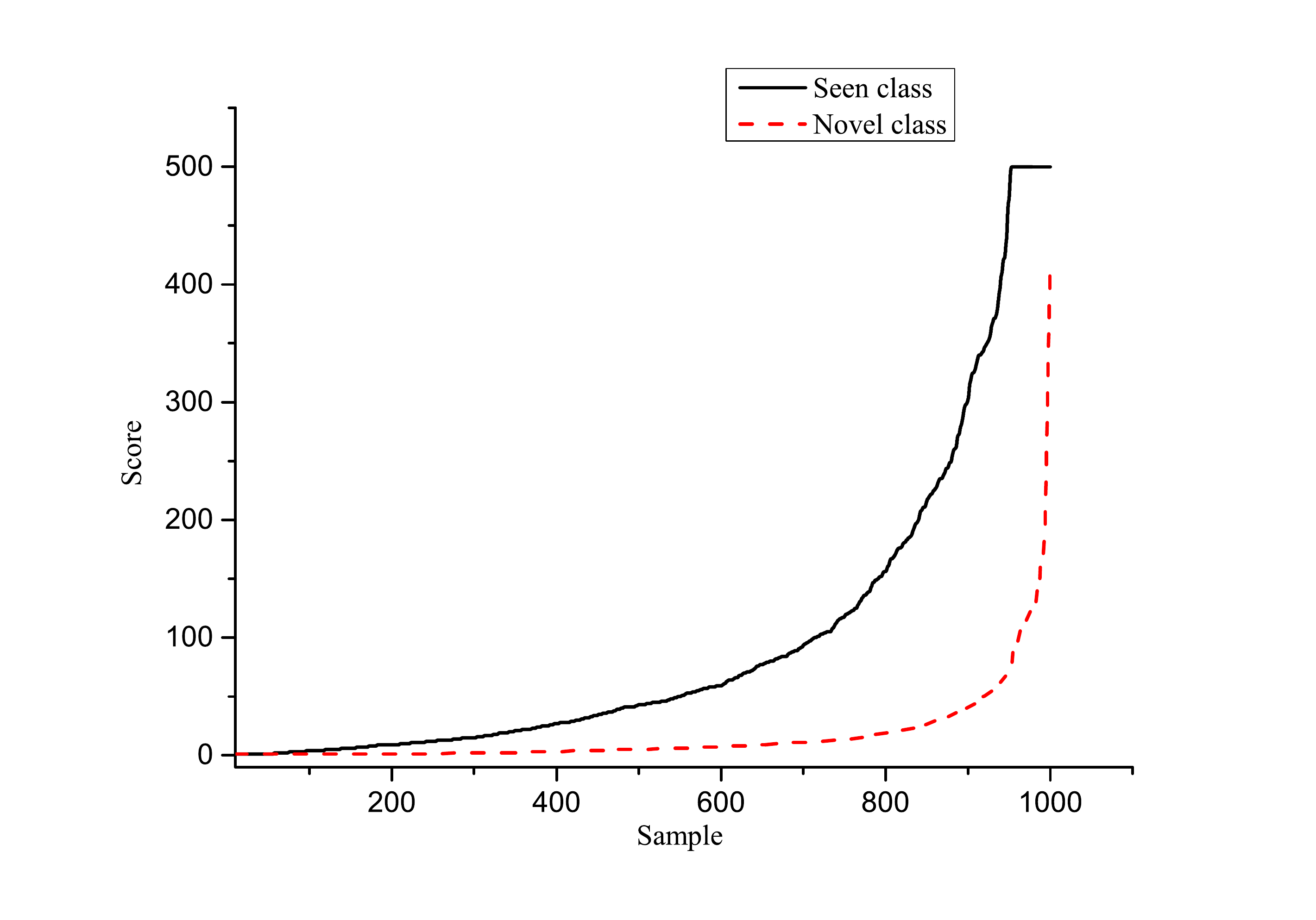}
    \vspace{-3\baselineskip}
    \caption{Visualization of differences in novelty score for known and novel classes}
    \label{fig:novelty_score}
\end{figure}

\begin{table}[ht]
\begin{center}
\begin{tabular}{|c|c|c|c|c|c|c|c|}
\hline \bf Dataset & \bf DT & \bf KNN & \bf SVM & \bf LR & \bf NB & \bf MLP \\ \hline
20 Newsgroup & 72.5 \% & 82 \% & 78.0 \% &72.75 \% & 69.0 \% & \textbf{82.50 \% }\\ \hline
Spooky action author & 53.42 \% & 57.89 \% & 63.15 \% & 52.63 \% & {58.68 }\% & \textbf{63.15 \%} \\ \hline
CMU movie & 61.05 \% & {64.73 \%} & 62.10 \% & 55.00 \% & 58.94 \% & \textbf{68.68} \% \\ \hline
BBC sports & 84.21 \% & {85.96 \%} & 75.43\% & 70.17 \% & 73.68 \% & \textbf{89.47} \%  \\ \hline
Web of Science & 64.70 \% & {67.97 \%} & 69.93 \% & 67.10 \% & 62.09 \% & \textbf{70.37} \% \\
\hline
\end{tabular}
\end{center}
\caption{\label{font-table} Accuracy of different machine learning classifiers to detect novel class in various dataset. }
\end{table}

The experimental results for all datasets are shown in Table \ref{font-table}. As seen, multilayer perceptron (MLP) (hidden layer sizes \textit{100,30} and relu activation with stochastic gradient descent) is superior for all of the datasets. In our experiments, the 20 Newsgroups and BBC sports datasets yielded better results than the three other data sets, arguably because of the sharp distinctiveness of examples in the known and novel classes. We further believe that the novelty scores are clustered based on known and novel classes, thus, distance-based methods seem effective in classification. We plotted the novelty score of thousand text samples from known and novel classes using our framework, which shows how scores differ significantly for each sample as visualized in Figure 3. Moreover, we can see from the graph that when the texts from the novel and known classes are discriminative, the TM produces distinct novelty scores, thus improving the ability of machine learning classifiers to detect novel texts. In all brevity, we believe that the clauses of a TM capture frequent patterns in the training data, and thus novel samples will reveal themselves by not fitting with a sufficiently large number of clauses. The number of triggered clauses can therefore be utilized to measure novelty score. Also, the clauses formed by trained TM models should only to a small degree trigger on novel classes, producing distinctively low scores.

\begin{table}[ht]
\begin{center}
\begin{tabular}{|c|c|c|c|c|c|}
\hline \bf Algorithms & \bf 20 Newsgroup & \bf Spooky action author & \bf CMU movie & \bf BBC sports & \bf WOS \\ \hline
LOF & 52.51 \% & 50.66 \% & 48.84 \% & 47.97 \% & 55.61 \%\\ \hline
Feature Bagging & 67.60 \% & 62.70 \% & 64.73 \% & 54.38 \% & 69.64 \% \\ \hline
HBOS & 55.03 \% & 48.55 \% & 48.57 \% & 49.53 \% & {55.09 \%} \\ \hline
Isolation Forest & 52.01 \% & 48.66\% & 49.10 \% & 49.35\% & 54.70 \% \\ \hline
Average KNN & 76.35 \% & 57.76 \% & 56.21 \% & 55.54 \% & \textbf{79.22} \% \\ \hline
K-Means clustering & 81.00 \% & 61.30 \% & 49.20 \% & 47.70 \% & 41.31 \% \\ \hline
One-class SVM  & \textbf{83.70 \%} &  43.56 \% & 51.94 \% & 83.53 \% & 36.32 \% \\ \hline
TM framework & 82.50 \% & \textbf{63.15\%} & \textbf{68.15 \%} & \textbf{89.47 \%} & 70.37 \% \\
\hline
\end{tabular}
\end{center}
\caption{\label{performance} Performance comparison of proposed TM framework with cluster and outlier based novelty detection algorithms. }
\end{table}

We compared the performance of our TM framework with different clustering and outlier detection algorithms such as Cluster-based Local Outlier Factor (CBLOF), Feature Bagging ($neighbors,\\n = 35$), Histogram-base Outlier Detection (HBOS), Isolation Forest, Average KNN, K-Means clustering, and One-class SVM. The evaluation was performed on the same preprocessed datasets for a fair comparison. To make comparision more robust, we preprocessed the data for the baseline algorithms using count vectorizer, term frequency-inverse document frequency (TFIDF) and Principle component analysis (PCA). Additionally, we utilized maximum possible outlier fraction (i.e., $0.5$) for these methods. The performance comparison is given in Table \ref{performance}, which shows that our framework surpasses the other algorithms on three of the datasets and performs competitively in the remaining two. However, in datasets like $WOS$, where there are many similar words shared between known and novel classes, our method is surprisingly surpassed by the distance-based algorithm. One-class SVM closely follows the performance of our TM framework, which may be due to its linear structure that prevents overfitting on imbalanced and small datasets. In our paper, we are not just trying to build a state-of-art novelty detection algorithm.

\section{Conclusions}
In this paper, we studied a problem of novelty detection in multiclass text classification. We proposed a score-based TM framework for novel class detection. We first used the clauses of the TM to produce a novelty score, distinguishing between known and novel classes. Then, a machine learning classifier is adopted for novelty classification using the novelty scores provided by the TM. The experimental results on various datasets demonstrate the effectiveness of our proposed framework. Our future work includes using a large text corpus with multiple classes for experimentation and studying the properties of the novelty score theoretically. 
%


\bibliographystyle{coling.bst}
\bibliography{coling2020.bib}
\end{document}